\documentclass{article}

\usepackage{microtype}
\usepackage{graphicx}
\usepackage{subfigure}
\usepackage{booktabs} 

\usepackage{hyperref}

\usepackage[accepted]{icml2024_tf2m}

\usepackage{amsmath}
\usepackage{amssymb}
\usepackage{mathtools}
\usepackage{amsthm}

\usepackage[capitalize,noabbrev]{cleveref}

\theoremstyle{plain}

\theoremstyle{definition}

\theoremstyle{remark}

\usepackage[textsize=tiny]{todonotes}

\icmltitlerunning{Attention Is All You Need But You Don't Need All Of It}

\begin{document}

\twocolumn[
\icmltitle{Attention Is All You Need But You Don't Need All Of It\\ For Inference of Large Language Models}



\icmlsetsymbol{equal}{*}

\begin{icmlauthorlist}
\icmlauthor{Georgy Tyukin}{equal,yyy}
\icmlauthor{Gbetondji J-S Dovonon}{yyy}
\icmlauthor{Jean Kaddour}{yyy}
\icmlauthor{Pasquale Minervini}{zzz}
\end{icmlauthorlist}

\icmlaffiliation{yyy}{University College London, UK}
\icmlaffiliation{zzz}{University of Edinburgh, UK}

\icmlcorrespondingauthor{Georgy Tyukin}{tyukinegor@gmail.com}

\icmlkeywords{Machine Learning, ICML}

\vskip 0.3in
]



\printAffiliationsAndNotice{\icmlEqualContribution} 

\begin{abstract}
The inference demand for LLMs has skyrocketed in recent months, and serving models with low latencies remains challenging due to the quadratic input length complexity of the attention layers.
In this work, we investigate the effect of dropping MLP and attention layers at inference time on the performance of Llama-v2 models.
We find that dropping dreeper attention layers only marginally decreases performance but leads to the best speedups alongside dropping entire layers.
For example, removing 33\% of attention layers in a 13B Llama2 model results in a 1.8\% drop in average performance over the OpenLLM benchmark.
We also observe that skipping layers except the latter layers reduces performances for more layers skipped, except for skipping the attention layers.
\end{abstract}

\section{Introduction}
The ubiquitous deployment of Large Language Models (LLMs) results in ever-growing amounts of compute spent on inference \citep{DBLP:journals/corr/abs-2104-10350,DBLP:journals/corr/abs-2302-01318,DBLP:journals/corr/abs-2307-10169,xia2024unlocking,reid2024gemini}.
Further, serving models with low latencies remains challenging because contemporary Transformer architectures employ the self-attention mechanism with quadratic input complexity \cite{touvron2023llama2,jiang2023mistral,bi2024deepseek}.
In this work, we delve deeper into the concept of layer skipping \cite{fan2019reducing,wang2022skipbert} to reduce the computation on superfluous LLM components.
Our findings demonstrate that pruning deeper attention layers does not significantly affect performance.
When applied to Llama-v2 \cite{touvron2023llama2}, we maintain good performance on the OpenLLM (ARC \cite{clark2018think}, HellaSwag \cite{zellers2019hellaswag}, MMLU
\cite{hendrycks2021measuring}, TruthfulQA \cite{lin2022truthfulqa}) benchmarks \cite{OpenLLMLeaderboardb}, recording only minimal performance deviations compared to the full model.

\section{Method}

\begin{figure}[ht]
    \centering
    \includegraphics[width=\linewidth]{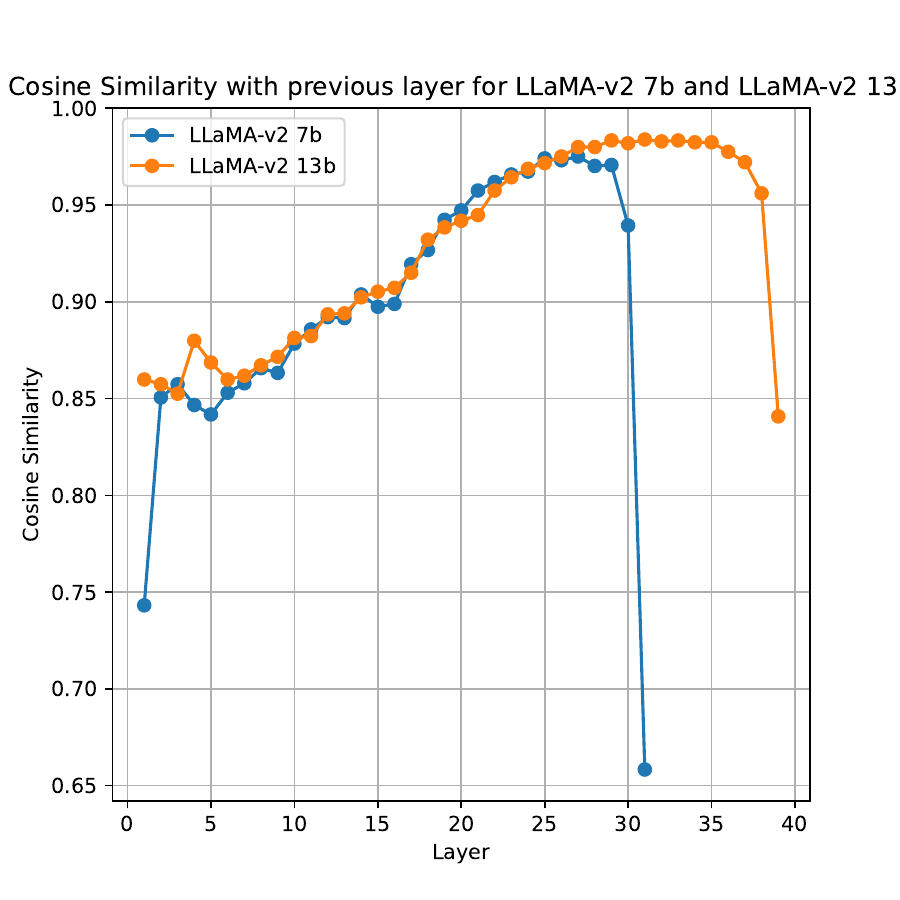}
    \caption{Cosine similarity of Llama-v2 layers with the previous layer: We observe that the deeper the layer, the more its features are similar to the previous layer except for the very last layer.}
    \label{fig:llama_cosine_similarity}
\end{figure}

\subsection{Layer skipping} Consider a Transformer model \(\mathcal{M}\) with \(L\) layers, each consisting of an attention sub-layer followed by a multi-layer perceptron (MLP) sub-layer. We denote each layer as \(\mathcal{M}_i = (\text{Attention}_i, \text{MLP}_i)\) for \(i \in \{1, 2, \ldots, L\}\).

To compare the performance of Transformer models when skipping specific sub-layers, we create two variants of the model:

1. \textbf{Skipping MLP Layers:} We construct a model \(\mathcal{M}_{\text{skip MLP}}\) by skipping the MLP sub-layer from the last \(k\) layers. The resulting model is \(\mathcal{M}_{\text{skip MLP}} = \{(\text{Attention}_i, \text{MLP}_i) \mid i \in \{1, 2, \ldots, L-k\}\} \cup \{(\text{Attention}_i, \emptyset) \mid i \in \{L-k+1, \ldots, L\}\}\).

2. \textbf{Skipping Attention Layers:} We construct a model \(\mathcal{M}_{\text{skip Attention}}\) by skipping the attention sub-layer from the last \(k\) layers. The resulting model is \(\mathcal{M}_{\text{skip Attention}} = \{(\text{Attention}_i, \text{MLP}_i) \mid i \in \{1, 2, \ldots, L-k\}\} \cup \{(\emptyset, \text{MLP}_i) \mid i \in \{L-k+1, \ldots, L\}\}\).

3. \textbf{Skipping Transformer Blocks:} We construct a model \(\mathcal{M}_{\text{skip Attention}}\) by skipping the entire last \(k\) layers. The resulting model is \(\mathcal{M}_{\text{skip Block}} = \{(\text{Attention}_i, \text{MLP}_i) \mid i \in \{1, 2, \ldots, L-k\}\} \cup \{(\emptyset) \mid i \in \{L-k+1, \ldots, L\}\}\).

We then evaluate the performance of these modified models on the OpenLLM benchmark \cite{OpenLLMLeaderboardb}, comparing metrics such as accuracy, computational efficiency, and memory usage. This comparison helps in understanding the individual contributions of the attention and MLP sub-layers to the overall performance of the Transformer model.

\begin{figure}[h!]
    \centering
    \resizebox{\columnwidth}{!}{%
        \begin{tabular}{c c c}
            \subfigure[Skip attention layers.]{
                \includegraphics[width=0.15\textwidth]{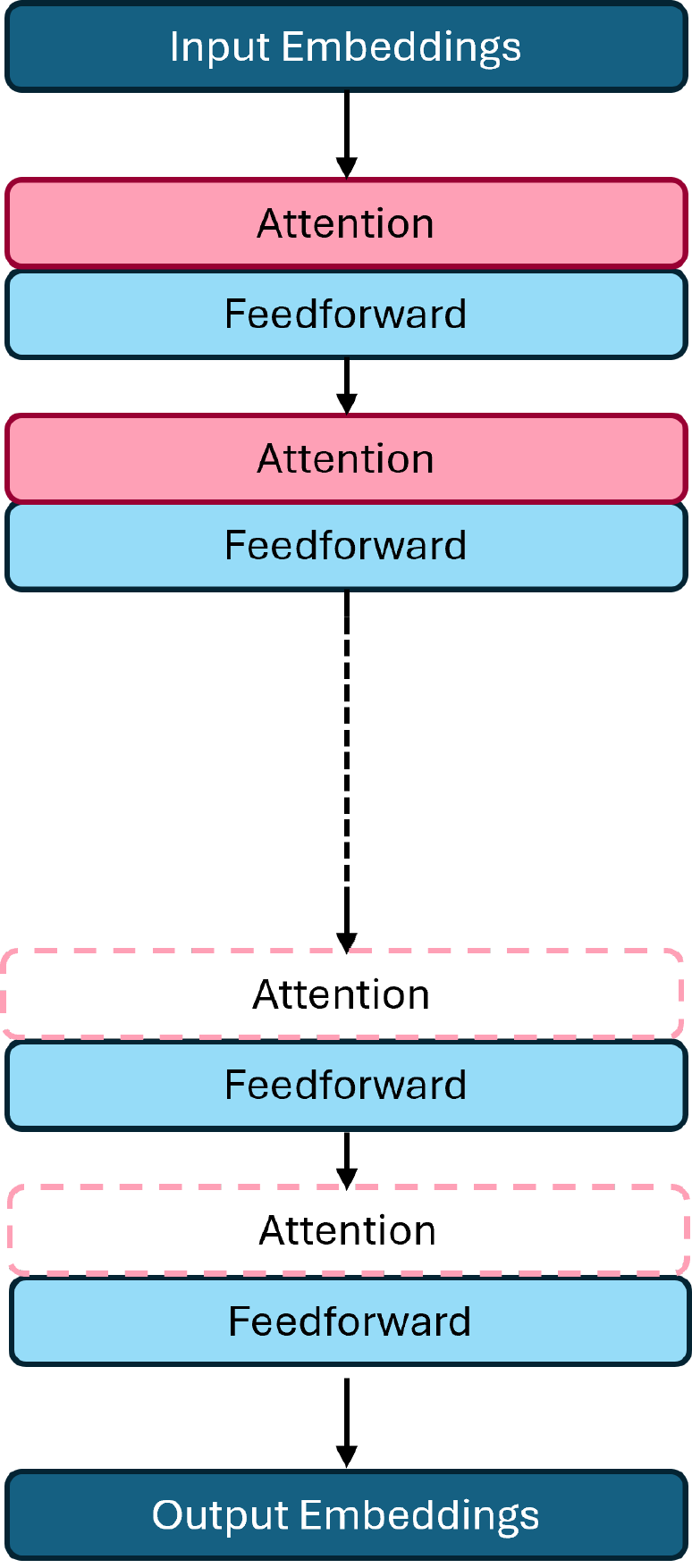}
            } &
            \subfigure[Skip attention layers, keep last full block.]{
                \includegraphics[width=0.15\textwidth]{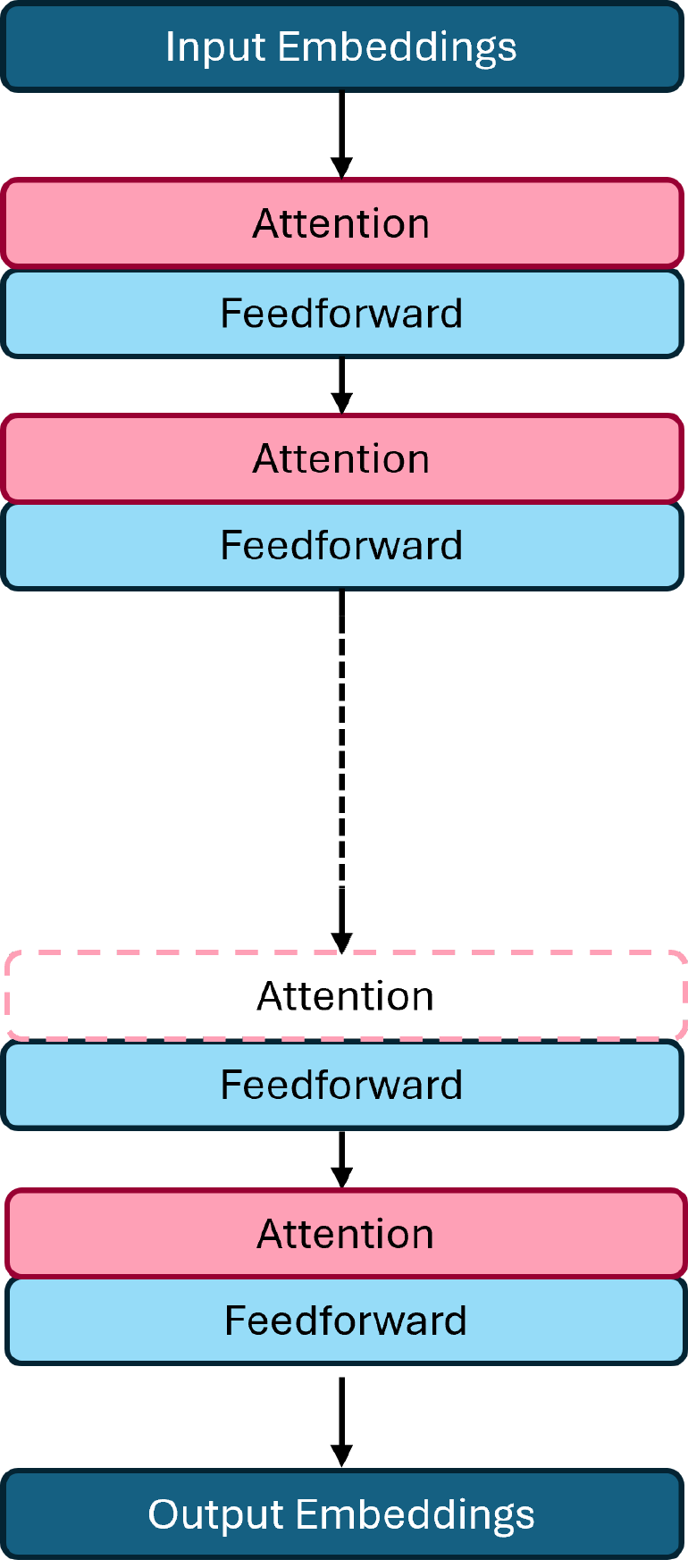}
            } &
            \subfigure[Skip ffwd layers.]{
                \includegraphics[width=0.15\textwidth]{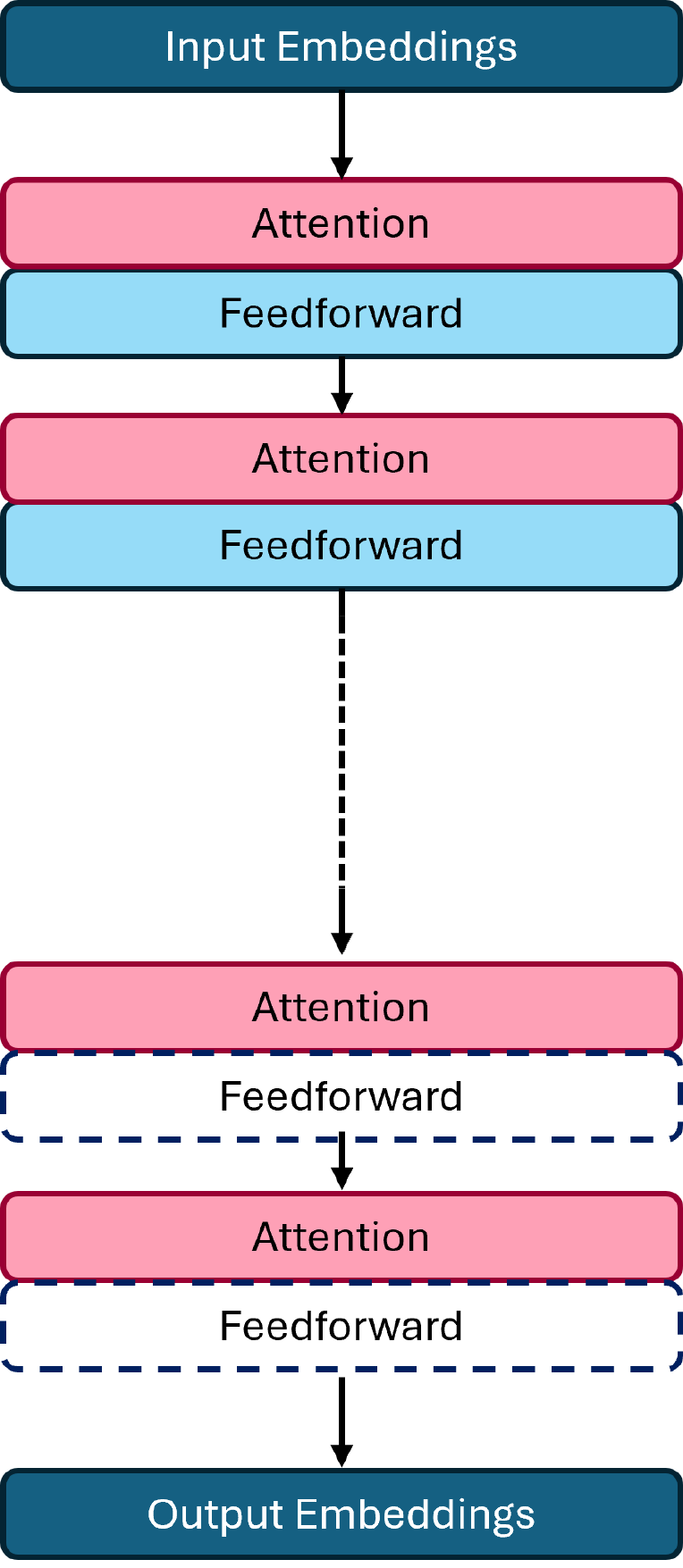}
            } \\
            \subfigure[Skip ffwd layers, keep last full block.]{
                \includegraphics[width=0.15\textwidth]{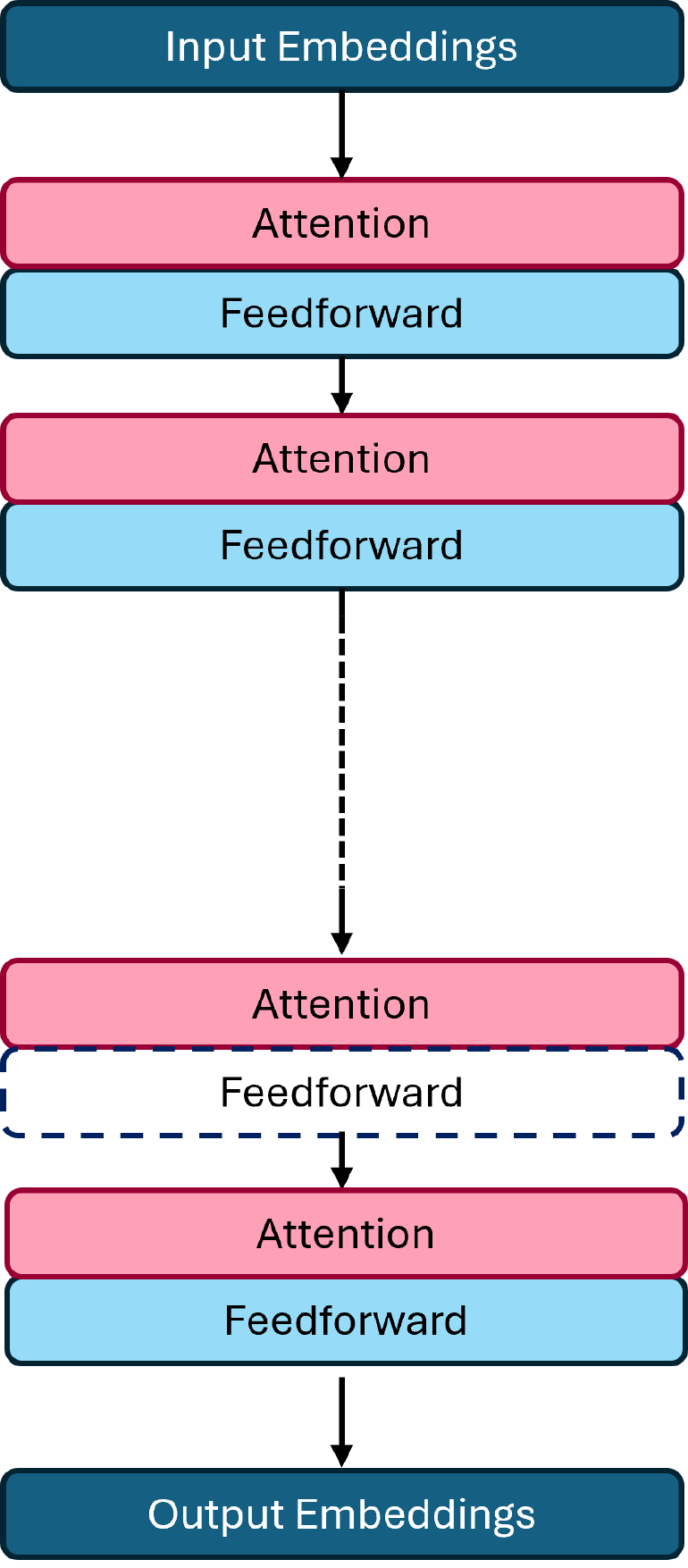}
            } &
            \subfigure[Skip full blocks.]{
                \includegraphics[width=0.15\textwidth]{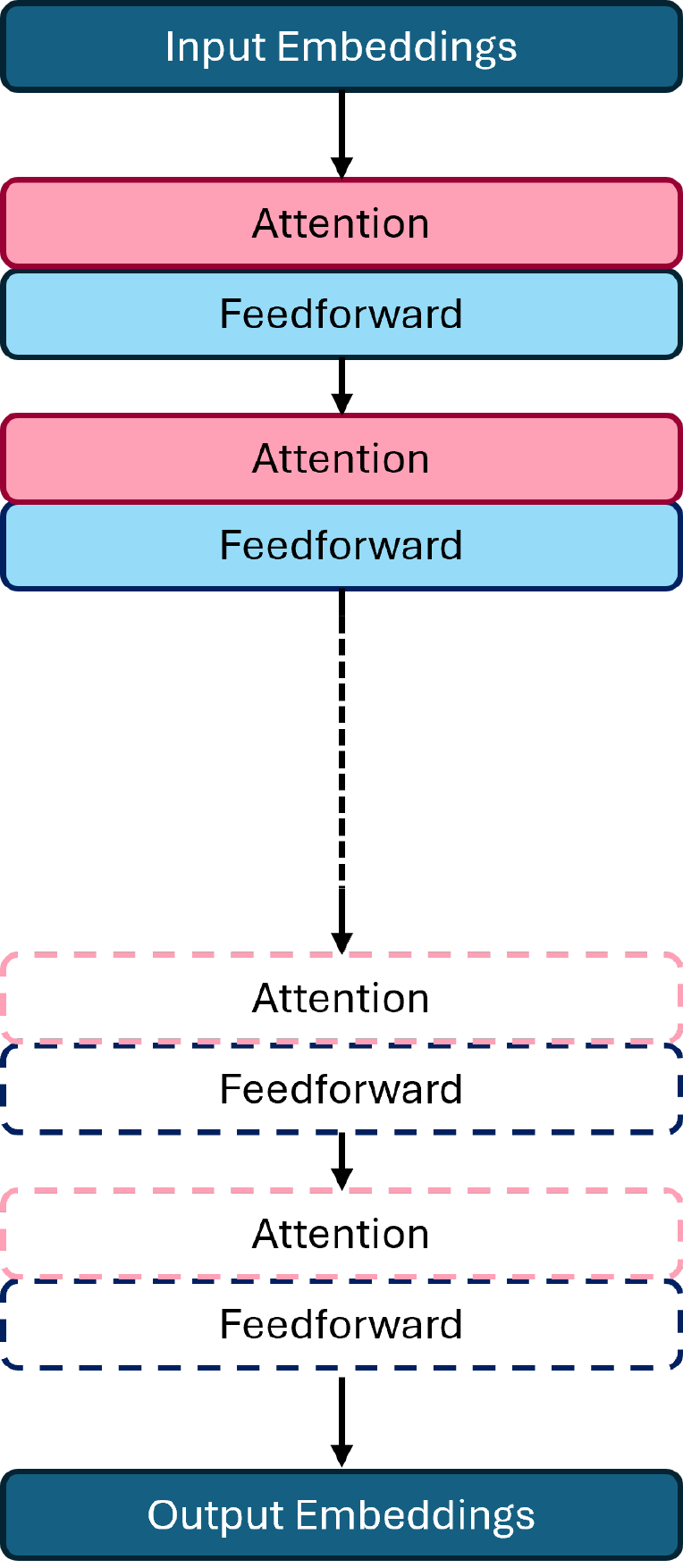}
            } &
            \subfigure[Skip full blocks, keep last full block.]{
                \includegraphics[width=0.15\textwidth]{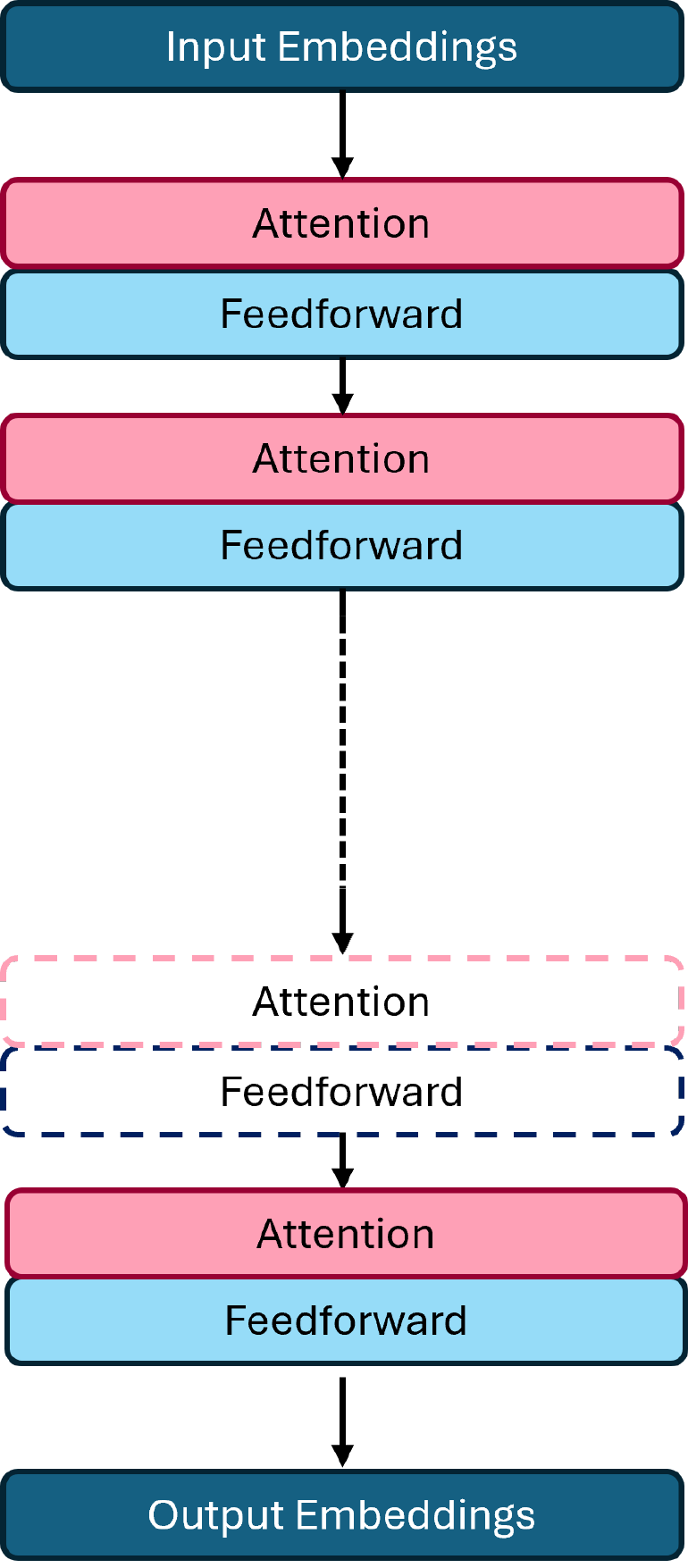}
            }
        \end{tabular}
    }
    \caption{\textbf{Skip mechanisms} for skipping single layers and entire Transformer blocks (ffwd and attention layers) during inference.}
    \label{fig:skip_mechanisms}
\end{figure}

\subsection{Motivation: Are Deeper Layers More Redundant?}

In Transformer models, the last layers have been shown to contribute less information than earlier layers, making it possible to drop those layers at a minimal performance cost \cite{fan2019reducing,zhang2020accelerating,wang2022skipbert,schuster2022confident,kaddour2023train,belrose2023eliciting}. 

To verify this, we experiment with removing either the attention sublayers or the MLP sublayers.
\Cref{fig:llama_cosine_similarity} shows the cosine similarities between a layer's features and the previous layer showing that deeper layers have a lower impact on the features than earlier layers. One notable exception to this trend is that the last layer for both Llama-v2 7B and 13B has the lowest cosine similarity with the previous layer.

Previous analysis of the attention mechanism has shown that they can converge to the same value due to attention collapse \cite{zhai2023stabilizing} and token features that also converge to the same value due to over-smoothing \cite{wang2022antioversmoothing,dovonon2024setting} or rank collapse \cite{dong2023attention}, with solutions to these issues typically improving performance \cite{ali2023centered,choi2024graph}. 

\section{Results}

\paragraph{Experimental Setup} 

For all experiments, we use either Llama-v2-7B or Llama-v2-13B \citep{touvron2023llama,touvron2023llama2}, two LLMs trained on trillions of publically available tokens. We experiment with keeping $66\%$, $75\%$, $90\%$ and $100\%$ of the network and report the corresponding results in \Cref{tab:main_results}. We also experiment with removing attention sublayers in \Cref{tab:main_results2}, MLP sublayers in \Cref{tab:main_results3}, and a varying number of layers similar to \Cref{tab:main_results} but keeping the last layer in \Cref{tab:main_results4}.

\subsection{Chopping Layers}


\begin{table}[h!]
\caption{Llama-v2 skipping full layer }
\label{tab:main_results}
\resizebox{0.93\linewidth}{!}{%
\begin{tabular}{@{}l|llll|l@{}}
\toprule
\multicolumn{1}{c|}{\textbf{Model}} &  \multicolumn{5}{c}{\textbf{Performances}}     \\ 
\cmidrule(l){2-6} 
\multicolumn{1}{c|}{}                                & ARC & HellaSwag & TruthfulQA & MMLU & \textbf{Average} \\ \midrule
7B-66\%  & 35.2 & 46.8 & 46.2 & 40.3 & \textbf{42.1} \\
7B-75\%  & 38.3 & 53.0 & 45.1 & 45.9 & \textbf{45.6} \\
7B-90\%   & 47.7 &  69.3 & 39.6 & 46.4 & \textbf{50.8} \\
7B-100\%  & 53.1 & 78.6 & 38.8 & 46.6 & \textbf{54.3} \\ \midrule
13B-66\%  & 37.8 & 46.8 & 45.3 & 51.8 & \textbf{45.4} \\
13B-75\%  & 40.9 & 53.6 & 42.5 & 53.2 & \textbf{47.6} \\
13B-90\%  & 51.3 & 71.3 & 37.1 & 54.8 & \textbf{53.6} \\
13B-100\% & 59.6 & 82.1 & 36.9 & 55.4 & \textbf{58.5} \\
\bottomrule
\end{tabular}%
}
\end{table}

\begin{table}[h!]
\caption{Llama-v2 skipping attention sublayers }
\label{tab:main_results2}
\resizebox{0.93\linewidth}{!}{%
\begin{tabular}{@{}l|llll|l@{}}
\toprule
\multicolumn{1}{c|}{\textbf{Model}} &  \multicolumn{5}{c}{\textbf{Performances}}     \\ 
\cmidrule(l){2-6} 
\multicolumn{1}{c|}{}                                & ARC & HellaSwag & TruthfulQA & MMLU & \textbf{Average} \\ \midrule
7B-66\%  & 51.2 & 77.0 & 42.2 & 39.4 & \textbf{52.5} \\
7B-75\%  & 52.5 & 78.3 & 42.3 & 41.4 & \textbf{53.6}\\
7B-90\%   & 52.8 &  78.9 & 40.0 & 44.0 & \textbf{53.9} \\
7B-100\%  & 53.1 & 78.6 & 38.8 & 46.6 & \textbf{54.3} \\ \midrule
13B-66\%  & 55.6 & 80.1 & 40.1 & 51.3 & \textbf{56.8} \\
13B-75\%  & 55.9 & 79.7 & 39.9 & 52.1 & \textbf{56.9} \\
13B-90\%  & 57.0 & 81.3 & 38.2 & 54.8 & \textbf{57.8} \\
13B-100\% & 59.6 & 82.1 & 36.9 & 55.4 & \textbf{58.5} \\
\bottomrule
\end{tabular}%
}
\end{table}

\begin{table}[h!]
\caption{Llama-v2 skipping ffwd sublayers }
\label{tab:main_results3}
\resizebox{0.93\linewidth}{!}{%
\begin{tabular}{@{}l|llll|l@{}}
\toprule
\multicolumn{1}{c|}{\textbf{Model}} &  \multicolumn{5}{c}{\textbf{Performances}}     \\ 
\cmidrule(l){2-6} 
\multicolumn{1}{c|}{}                                & ARC & HellaSwag & TruthfulQA & MMLU & \textbf{Average} \\ \midrule
7B-66\%  & 35.1 & 52.5 & 42.2 & 43.9 & \textbf{43.4} \\
7B-75\%  & 40.4 & 60.3 & 39.2 & 46.3 & \textbf{46.6}\\
7B-90\%  & 48.5 & 71.4 & 38.0 & 46.1 & \textbf{51.0} \\
7B-100\% & 53.1 & 78.6 & 38.8 & 46.6 & \textbf{54.3} \\ \midrule
13B-66\%  & 41.6 & 56.9 & 40.7 & 53.4 & \textbf{48.2} \\
13B-75\%  & 47.3 & 65.2 & 40.0 & 53.2 & \textbf{51.4} \\
13B-90\%  & 54.2 & 75.8 & 38.3 & 54.7 & \textbf{55.8} \\
13B-100\% & 59.6 & 82.1 & 36.9 & 55.4 & \textbf{58.5} \\
\bottomrule
\end{tabular}%
}
\end{table}

On all datasets except TruthfulQA, performance drops which is expected. It had already been observed that larger language models are less truthful \citep{lin2022truthfulqa}, but we now also observe that reducing the size of already trained models can also make them more truthful. The observation still holds when the last layer is preserved. 
Skipping attention layers only leads to better results with only a $1.8\%$ decrease in performance when keeping $66\%$ of the network compared to a $13.1\%$ decrease in performance when dropping dropping the MLP layers only. This seems to indicate that MLP layers are more important than attention layers, at least in deeper parts of the network.

\subsection{Last Layer Inclusion}

\begin{table}[h!]
\caption{Llama-v2 skip full layers with last layer }
\label{tab:main_results4}
\resizebox{0.93\linewidth}{!}{%
\begin{tabular}{@{}l|llll|l@{}}
\toprule
\multicolumn{1}{c|}{\textbf{Model}} &  \multicolumn{5}{c}{\textbf{Performances}}     \\ 
\cmidrule(l){2-6} 
\multicolumn{1}{c|}{}                                & ARC & HellaSwag & TruthfulQA & MMLU & \textbf{Average} \\ \midrule
7B-66\%  & 32.0 & 45.8 & 46.9 & 40.7 & \textbf{41.3} \\
7B-75\%  & 34.5 & 49.4 & 45.9 & 38.3 & \textbf{42.0}\\
7B-90\%  & 46.5 & 73.1 & 41.8 & 41.4 & \textbf{50.7} \\
7B-100\% & 53.1 & 78.6 & 38.8 & 46.6 & \textbf{54.3} \\ \midrule
13B-66\%  & 35.1 & 50.0 & 46.9 & 19.1 & \textbf{37.8} \\
13B-75\%  & 38.7 & 56.6 & 43.7 & 25.2& \textbf{41.1} \\
13B-90\%  & 51.2 & 78.1 & 38.0 & 27.1 & \textbf{47.9} \\
13B-100\% & 59.6 & 82.1 & 36.9 & 55.4 & \textbf{58.5} \\
\bottomrule
\end{tabular}%
}
\end{table}

\begin{table}[h!]
\caption{Llama-v2 skip attention sublayers with last layer }
\label{tab:main_results5}
\resizebox{0.93\linewidth}{!}{%
\begin{tabular}{@{}l|llll|l@{}}
\toprule
\multicolumn{1}{c|}{\textbf{Model}} &  \multicolumn{5}{c}{\textbf{Performances}}     \\ 
\cmidrule(l){2-6} 
\multicolumn{1}{c|}{}                                & ARC & HellaSwag & TruthfulQA & MMLU & \textbf{Average} \\ \midrule
7B-66\%  & 49.3 & 77.1 & 40.5 & 42.5 & \textbf{52.4} \\
7B-75\%  & 51.8 & 78.3 & 41.1 & 44.1 & \textbf{53.8} \\
7B-90\%   & 51.9 &  78.7 & 39.4 & 45.7 & \textbf{53.9} \\
7B-100\%  & 53.1 & 78.6 & 38.8 & 46.6 & \textbf{54.3} \\ \midrule
13B-66\%  & 56.8 & 82.1 & 38.0 & 50.3 & \textbf{56.8} \\
13B-75\%  & 57.5 & 82.1 & 37.0 & 51.4 & \textbf{57.0} \\
13B-90\%  & 58.9 & 82.4 & 36.6 & 54.5 & \textbf{58.1} \\
13B-100\% & 59.6 & 82.1 & 36.9 & 55.4 & \textbf{58.5} \\
\bottomrule
\end{tabular}%
}
\end{table}

\begin{table}[h!]
\caption{Llama-v2 skip ffwd sublayers with last layer }
\label{tab:main_results6}
\resizebox{0.93\linewidth}{!}{%
\begin{tabular}{@{}l|llll|l@{}}
\toprule
\multicolumn{1}{c|}{\textbf{Model}} &  \multicolumn{5}{c}{\textbf{Performances}}     \\ 
\cmidrule(l){2-6} 
\multicolumn{1}{c|}{} & ARC & HellaSwag & TruthfulQA & MMLU & \textbf{Average} \\ \midrule
7B-66\%  & 32.0 & 45.8 & 46.9 & 39.4 & \textbf{41.0}\\
7B-75\%  & 34.5 & 49.4 & 45.9 & 40.2 & \textbf{42.5}\\
7B-90\%   & 46.5 &  73.1 & 41.8 & 40.2 & \textbf{50.4} \\
7B-100\%  & 53.1 & 78.6 & 38.8 & 46.6 & \textbf{54.3} \\ \midrule
13B-66\%  & 35.1 & 50.0 & 46.9 & 20.4 & \textbf{38.1} \\
13B-75\%  & 38.7 & 56.6 & 43.7 & 33.6 & \textbf{43.2} \\
13B-90\%  & 51.2 & 78.1 & 38.0 & 34.4 & \textbf{50.4} \\
13B-100\% & 59.6 & 82.1 & 36.9 & 55.4 & \textbf{58.5} \\
\bottomrule
\end{tabular}%
}
\end{table}

Surprisingly, we notice that skipping layers except the latter layers reduces performances for more layers skipped, except for skipping the attention layers. This is even more exaggerated compared to just dropping layers, including the last one. The reason for this could be attributed to the (lack of) robustness of feedforward sublayers, as the last layer now has to process perturbed information from earlier layers. For future work, it would be interesting to see if these performance drops can be compensated by a small amount of continued training; since model growing techniques for training seem to not suffer from instabilities \cite{kaddour2023train}.

\subsection{Compute-matched Comparison}
To measure the efficiency of the networks we conducted a separate experiment, where we record the time it takes for the model to output a sequence of length 1, averaging over 1000 sequences. We conducted this experiment for both 50 and 100 length input sequences. We notice that full layer droppings do improve time costs the best, followed by attention sublayers, and then feedforward sublayers which do not impact the speed of processing a lot.

We report the time$\times 10^2$ (for clarity) it takes to predict 1 token for 1000 sequences  as well as the percentage improvement. We show the results of this experiment for Llama 2 7B with 0\%, 10\%, 25\%, 33\% of layers skipped and we label these as 7B-100\%, 7B-90\%, 7B-75\%, 7B-66\% respectively.

\begin{table}[h!]
\caption{Llama-v2 time results, 50 length sequence, no last layer}
\label{tab:time_results}
\resizebox{\linewidth}{!}{%
\begin{tabular}{@{}l|ll|ll|ll@{}}
\toprule
\multicolumn{1}{c|}{\textbf{Model}} &  \multicolumn{2}{c}{\textbf{Full}} & \multicolumn{2}{c}{\textbf{Attention}} & \multicolumn{2}{c}{\textbf{ffwd}}   \\ 
\cmidrule(l){2-7}
\multicolumn{1}{c|}{} & Time(s) $\times 10^2$ & (\%) & Time(s) $\times 10^2$ & (\%) & Time(s) $\times 10^2$ & (\%)\\ \midrule
7B-66\%  & 31.35 & 32.96 & 36.72 & 21.47 &  43.51 & 6.95 \\
7B-75\%  & 35.48 & 24.12 & 39.46 & 15.61 & 42.88 & 8.30\\
7B-90\%   & 43.31 & 7.38 & 42.93 & 8.19 & 44.17 & 5.53 \\
\midrule
7B-100\%  & 46.76 & 0 & - & - & - & - \\ 
\bottomrule
\end{tabular}%
}
\end{table}

\begin{table}[h!]
\caption{Llama-v2 time results, 50 length sequence, last layer included}
\label{tab:time_results2}
\resizebox{\linewidth}{!}{%
\begin{tabular}{@{}l|ll|ll|ll@{}}
\toprule
\multicolumn{1}{c|}{\textbf{Model}} &  \multicolumn{2}{c}{\textbf{Full}} & \multicolumn{2}{c}{\textbf{Attention}} & \multicolumn{2}{c}{\textbf{ffwd}}   \\ 
\cmidrule(l){2-7}
\multicolumn{1}{c|}{} & Time(s) $\times 10^2$ & (\%) & Time(s) $\times 10^2$ & (\%) & Time(s) $\times 10^2$ & (\%)\\ \midrule
7B-66\%  & 31.78 & 32.04 & 36.92 & 21.04 &  41.31 & 11.66\\
7B-75\%  & 34.98 & 25.19 & 40.24 & 13.94 & 42.62 & 8.85\\
7B-90\%   & 40.92 & 12.49 & 42.43 & 9.26 & 43.51 & 6.95  \\
\midrule
7B-100\%  & 46.76 & 0 & - & - & - & - \\ 
\bottomrule
\end{tabular}%
}
\end{table}

\begin{table}[h!]
\caption{Llama-v2 time results, 100 length sequence, no last layer}
\label{tab:time_results3}
\resizebox{\linewidth}{!}{%
\begin{tabular}{@{}l|ll|ll|ll@{}}
\toprule
\multicolumn{1}{c|}{\textbf{Model}} &  \multicolumn{2}{c}{\textbf{Full}} & \multicolumn{2}{c}{\textbf{Attention}} & \multicolumn{2}{c}{\textbf{ffwd}}   \\ 
\cmidrule(l){2-7}
\multicolumn{1}{c|}{} & Time(s) $\times 10^2$ & (\%) & Time(s) $\times 10^2$ & (\%) & Time(s) $\times 10^2$ & (\%)\\ \midrule
7B-66\%  & 32.36 & 32.58 & 38.97 & 18.18 &  43.08 & 10.25\\
7B-75\%  & 36.58 & 23.79 & 41.27 & 14.02 & 44.13 & 8.06 \\
7B-90\%   & 43.65 & 9.06 & 44.62 & 7.04 & 46.30 & 3.54 \\
\midrule
7B-100\%  & 48.00 & 0 & - & - & - & - \\ 
\bottomrule
\end{tabular}%
}
\end{table}

\begin{table}[h!]
\caption{Llama-v2 time results, 100 length sequence, last layer included}
\label{tab:time_results4}
\resizebox{\linewidth}{!}{%
\begin{tabular}{@{}l|ll|ll|ll@{}}
\toprule
\multicolumn{1}{c|}{\textbf{Model}} &  \multicolumn{2}{c}{\textbf{Full}} & \multicolumn{2}{c}{\textbf{Attention}} & \multicolumn{2}{c}{\textbf{ffwd}}   \\ 
\cmidrule(l){2-7}
\multicolumn{1}{c|}{} & Time(s) $\times 10^2$ & (\%) & Time(s) $\times 10^2$ & (\%) & Time(s) $\times 10^2$ & (\%)\\ \midrule
7B-66\%  & 32.05 &33.23  & 38.52 & 19.75 &  42.66 & 11.13\\
7B-75\%  & 36.41 & 24.15 & 41.00 & 14.58 & 43.92 & 8.50\\
7B-90\%   & 43.28 & 9.83 & 44.27 & 7.77 & 45.20 & 5.83 \\
\midrule
7B-100\%  & 48.00 & 0 & - & - & - & - \\ 
\bottomrule
\end{tabular}%
}
\end{table}

\section{Related Work}

\paragraph{Early Exit during inference} Early exit methods have also been proposed in other domains \citep{graves2017adaptive,teerapittayanon2017branchynet} before getting adapted to autoregressive models \citep{Elbayad2020Depth-Adaptive,schuster2022confident,din2023jump,elhoushi2024layer,fan2024layers,chen2024eellm}. The idea works by dynamically allocating compute based on the difficulty of the input sequence. Our method prunes the deepest layers and does not involve any level of adaptability. This is beneficial because it does not require the entire model to be loaded in memory.
Dropping layers during inference has been done on BERT-like models in \citep{wang2022skipbert,Sajjad_2023}. We apply a similar analysis to more recent LLMs and study the impact of skipping attention and/or MLP layers in more detail. Concurrent work to ours by \citet{gromov2024unreasonable} yields similar results by pruning deeper layers and applying fine-tuning on the pruned model.  

\paragraph{Layer dropping/growing during training} There are various works studying the dropping/growing layers dynamically during training \citep{fan2019reducing,gong2019efficient,kaddour2023train,9180094,pmlr-v202-liu23am}. In contrast, this work focuses on dropping layers of an already pre-trained model in a way similar to \citet{men2024shortgptlayerslargelanguage}.

\paragraph{Other Inference Speedup Methods} Other works to speed up inference include compressing KV caches \cite{nawrot2024dynamic,wu2024layercondensed,bi2024deepseek}, speculative decoding \cite{DBLP:journals/corr/abs-2302-01318}, efficient memory management \cite{kwon2023efficient}, or subqudratic attention architectures \cite{fu2022hungry,peng2023rwkv,gu2023mamba}, an overview has been provided by \citet{DBLP:journals/corr/abs-2307-10169}.


\section{Conclusion}

We investigated the effect of dropping the last layers from the 7B and 13B Llama2 models. We observe that dropping attention sublayers lead to much lower drops in performance than dropping the MLP sublayers, whether the last layer is included or not, while also leading to better inference speedups. For example, removing 33\% of attention layers leads to an 18\% speedup in a 13B Llama2 model at the cost of a 1.8\% drop in average performance. This shows that massive improvements can be made over dropping entire layers from just dropping the attention sublayer.

\bibliographystyle{icml2024}
\bibliography{refs}

\end{document}